\documentclass[subba]{imsart}
\pubyear{0000}
\volume{00}
\issue{0}
\arxiv{2002.10243}
\firstpage{1}
\lastpage{25}

\usepackage{amsthm}
\usepackage{amsmath}
\usepackage{amssymb}
\usepackage{natbib}
\usepackage{makecell}
\usepackage[colorlinks,citecolor=blue,urlcolor=blue,filecolor=blue,backref=page]{hyperref}
\usepackage{graphicx}
\newtheorem{theorem}{Theorem}

\DeclareMathOperator*{\argmin}{arg\,min}

\startlocaldefs
\endlocaldefs

\begin{document}


\begin{frontmatter}
\title{Informative Bayesian Neural Network Priors for Weak Signals}

\runtitle{Informative Neural Network Priors}

\begin{aug}
\author{\fnms{Tianyu} \snm{Cui}\thanksref{addr1}\ead[label=e1]{firstname.lastname@aalto.fi}},
\author{\fnms{Aki} \snm{Havulinna}\thanksref{addr2,addr3}\ead[label=e1]{firstname.lastname@thl.fi}},
\author{\fnms{Pekka} \snm{Marttinen}\thanksref{addr1,addr2,t1}\ead[label=e1]{firstname.lastname@aalto.fi}}
\and
\author{\fnms{Samuel} \snm{Kaski}\thanksref{addr1,addr4,t1}\ead[label=e1]{firstname.lastname@aalto.fi}}

\runauthor{T. Cui et al.}

\address[addr1]{Helsinki Institute for Information Technology HIIT, Department of Computer Science, Aalto University, Finland
\printead{e1}
}

\address[addr2]{Finnish Institute for Health and Welfare (THL), Finland
}
\address[addr3]{Institute for Molecular Medicine Finland, FIMM-HiLIFE, Helsinki, Finland}
\address[addr4]{Department of Computer Science, University of Manchester, UK}

\thankstext{t1}{Equal contribution}


\end{aug}

\begin{abstract}
Encoding domain knowledge into the prior over the high-dimensional weight space of a neural network is challenging but essential in applications with limited data and weak signals. Two types of domain knowledge are commonly available in scientific applications: 1. feature sparsity (fraction of features deemed relevant); 2. signal-to-noise ratio, quantified, for instance, as the proportion of variance explained (PVE). We show how to encode both types of domain knowledge into the widely used Gaussian scale mixture priors with Automatic Relevance Determination. Specifically, we propose a new joint prior over the local (i.e., feature-specific) scale parameters that encodes knowledge about feature sparsity, and a Stein gradient optimization to tune the hyperparameters in such a way that the distribution induced on the model's PVE matches the prior distribution. We show empirically that the new prior improves prediction accuracy, compared to existing neural network priors, on several publicly available datasets and in a genetics application where signals are weak and sparse, often outperforming even computationally intensive cross-validation for hyperparameter tuning.
\end{abstract}
\begin{keyword}
\kwd{Informative prior}
\kwd{Neural network}
\kwd{Proportion of variance explained}
\kwd{Sparsity}
\end{keyword}
\end{frontmatter}

\section{Introduction}

Neural networks (NNs) have achieved state-of-the-art performance on a wide range of supervised learning tasks with high a signal-to-noise ratio (S/N), such as computer vision \citep{krizhevsky2012imagenet} and natural language processing \citep{devlin2018bert}. However, NNs often fail in scientific applications where domain knowledge is essential, e.g., when data are limited or the signal is extremely weak and sparse. Applications in genetics often fall into the latter category and are used as the motivating example for our derivations. Bayesian approach \citep{gelman2013bayesian} has been of interest in the NN community because of its ability to incorporate domain knowledge into reasoning and to provide principled handling of uncertainty. Nevertheless, it is still largely an open question how to encode domain knowledge into the prior over Bayesian neural network (BNN) weights, which are often high-dimensional and uninterpretable. 
\begin{figure*}[t]
    \centering
    \includegraphics[width=1.0\linewidth]{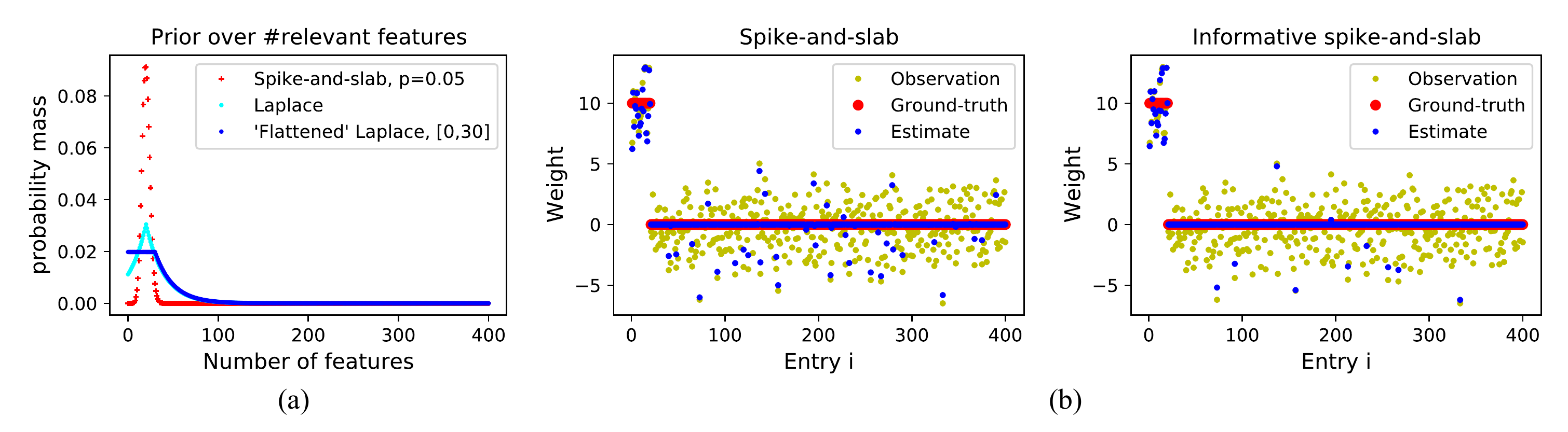}
    \caption{\textbf{a)} A spike-and-slab prior with slab probability $p=0.05$ induces a binomial distribution on the number of relevant features. The proposed informative spike-and-slab can encode a spectrum of alternative beliefs, such as a discretized or 'flattened' Laplace (for details, see Section 3). \textbf{b)} The informative spike-and-slab prior can remove false features more effectively than the vanilla spike-and-slab prior with correct slab probability, where features are assumed independent (see Section 6.1).}
    \label{fig: illustration_sparsity}
\end{figure*}
\begin{figure*}[t]
    \centering
    \includegraphics[width=1.0\linewidth]{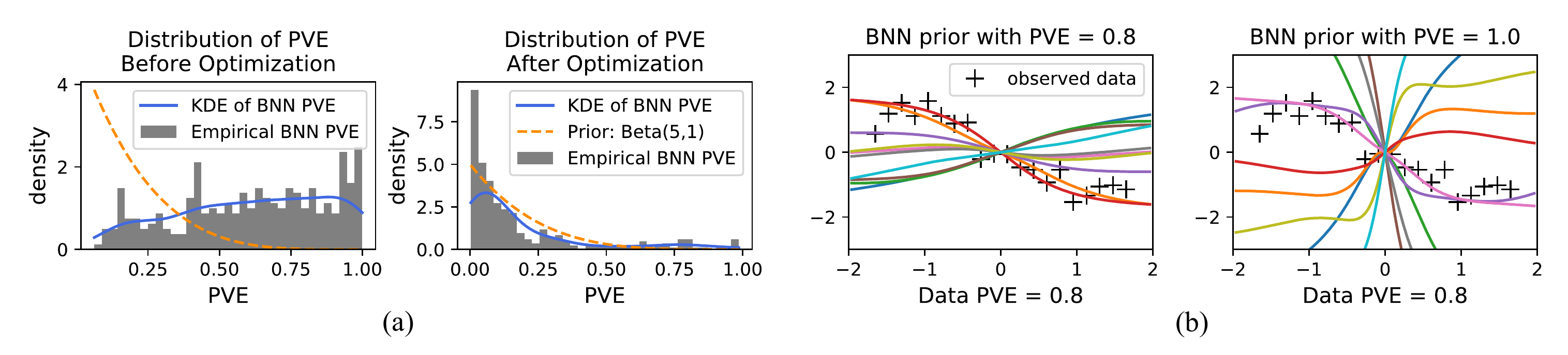}
    \caption{\textbf{a)} The empirical distribution and corresponding kernel density estimation (KDE) of the proportion of variance explained (PVE) for a BNN, obtained by simulating from the model, before and after optimizing the hyperparameters according to the prior belief on the PVE. \textbf{b)} The data with PVE=0.8 in its generating process are more likely to be generated by a BNN when the mode of the PVE is approximately correctly (left) than incorrectly (right). Colored lines are functions sampled from the BNN (for details, see Section 4).}
    \label{fig: illustration_pve}
\end{figure*}

We study the family of Gaussian scale mixture (GSM) \citep{andrews1974scale} distributions, which are widely used as priors for BNN weights. A particular example of interest is the spike-and-slab prior \citep{mitchell1988bayesian}
\begin{equation}
\begin{split}
w_{i,j}^{(l)}|\sigma, \lambda_i^{(l)}, \tau_i^{(l)} &\sim \mathcal{N}(0, \sigma^{(l)2}\lambda_{i}^{(l)2}\tau_i^{(l)2}), \;\; \tau_i^{(l)}\sim \text{Bernoulli}(p),
\label{eq: spike-and-slab 1}
\end{split}
\end{equation}
where $w_{i,j}^{(l)}$ represents the NN weight from node $i$ in layer $l$ to node $j$ in layer $l+1$. The hyper-parameters $\{\sigma^{(l)}, \lambda_{i}^{(l)}, p\}$ are often given non-informative hyper-priors \citep{neal2012bayesian}, such as the inverse Gamma on $\sigma^{(l)}$ and $\lambda_i^{(l)}$, or optimized using cross-validation \citep{blundell2015weight}. In contrast, we propose determining the hyper-priors according to two types of domain knowledge often available in scientific applications: ballpark figures on feature sparsity and the signal-to-noise ratio. Feature sparsity refers to the expected fraction of features used by the model. For example, it is known that less than 2\% of the genome encodes for genes, which may inform the expectation on the fraction of relevant features in a genetics application. A prior on the signal-to-noise ratio specifies the amount of target variance expected to be explained by the chosen features, and it can be quantified as the proportion of variance explained (PVE) \citep{glantz1990primer}. For instance, one gene may explain a tiny fraction of the variance of a given phenotype (prediction target in genetics, e.g. the height of an individual), i.e., the PVE of a gene may be as little as 1\%. 

Existing scalable sparsity-inducing BNN priors, such as the spike-and-slab prior, are restricted in the forms of prior beliefs about sparsity they can express: conditionally on the slab probability $p$ the number of relevant features follows a Binomial distribution. Specifying a Beta hyper-prior on $p$ could increase flexibility, but this still is more restricted and less intuitive than specifying any distribution directly on the number of relevant features, and in practice in the BNN literature a point estimate for $p$ is used. The value of $p$ is either set manually, cross-validated, or optimized as part of MAP estimation \citep{deng2019adaptive}. Moreover, the weights for different features or nodes are (conditionally) independent in Equation \ref{eq: spike-and-slab 1}; thus, incorporating correct features will not help remove false ones. In this paper, we propose a novel informative hyper-prior over the feature inclusion indicators $\tau^{(l)}_{i}$, called informative spike-and-slab, which can directly model any distribution on the number of relevant features (Figure \ref{fig: illustration_sparsity}a). In addition, unlike the vanilla spike-and-slab, the $\tau^{(l)}_{i}$ for different features $i$ are dependent in the new informative spike-and-slab, and consequently false features are more likely to be removed when correct features are included, which can be extremely beneficial when the noise level is high, as demonstrated with a toy example in Figure \ref{fig: illustration_sparsity}b.

The PVE assumed by a BNN affects the variability of functions drawn from the prior (Figure \ref{fig: illustration_pve}b). Intuitively, when the PVE of a BNN is close to the correct PVE, the model is more likely to recover the underlying data generating function. The distribution of PVE assumed by a BNN is induced by the prior on the model's weights, which in turn is affected by all the hyper-parameters. Thus, hyper-parameters that do not affect feature sparsity, e.g. $\lambda_i^{(l)}$, can be used to encode domain knowledge about the PVE. We propose a scalable gradient-based optimization approach to match the model's PVE with the prior belief on the PVE, e.g., a Beta distribution, by minimizing the Kullback–Leibler divergence between the two distributions w.r.t. chosen hyper-parameters using the Stein gradient estimator \citep{li2018gradient} (Figure \ref{fig: illustration_pve}a). Although it has been demonstrated that using cross-validation to specify hyper-parameters, e.g. the global scale in the mean-field prior, is sufficient for tasks with a high S/N and a large dataset \citep{wilson2020bayesian}, we empirically show that being informative about the PVE can improve performance in low S/N and small data regimes, even without computationally intensive cross-validation.

The structure of this paper is the following. Section \ref{sec:background} reviews required background on Bayesian neural networks and Stein gradients. In Section \ref{sec: sparsity info}, we describe our novel joint hyper-prior over the local scales which explicitly encodes feature sparsity. In section \ref{sec: PVE}, we present the novel optimization algorithm to tune the distribution of a model's PVE according to prior knowledge. Section \ref{sec:related} reviews in detail a large body of related literature on BNNs. Thorough experiments with synthetic and real-world data sets are presented in Section \ref{sec:experiments}, demonstrating the benefits of the method. Finally, Section \ref{sec:conclusion} concludes, including discussion on limitations of our method as well as suggested future directions.

\section{Background}\label{sec:background}
\subsection{Proportion of Variance Explained}
In regression tasks, we assume that the data generating process takes the form
\begin{equation}
y = f(\mathbf{x};\mathbf{w})+\epsilon,
\label{eq: data generating model}
\end{equation}
where $f(\mathbf{x};\mathbf{w})$ is the unknown target function, and $\epsilon$ is the unexplainable noise. The Proportion of Variance Explained (PVE) \citep{glantz1990primer} of $f(\mathbf{x};\mathbf{w})$ on dataset $\{\mathbf{X},\mathbf{y}\}$ with input $\mathbf{X}=\{\mathbf{x}^{(1)},\ldots,\mathbf{x}^{(N)}\}$ and outputs $\mathbf{y}=\{y^{(1)},\ldots,y^{(N)}\}$,  also called the coefficient of determination ($R^2$) in linear regression, is
\begin{equation}
\begin{split}
\text{PVE}(\mathbf{w}) &= 1-\frac{\sum_{i}^{N}(y^{(i)}-f(\mathbf{x}^{(i)};\mathbf{w}))^2}{\sum_{i}^{N}(y^{(i)}-\bar{y})^2}
=1 - \frac{\text{Var}(\epsilon)}{\text{Var}(f(\mathbf{x}^{(i)};\mathbf{w}))+\text{Var}(\epsilon)}.
\label{eq: PVE}
\end{split}
\end{equation}
 The PVE is commonly used to measure the impact of features $\mathbf{x}$ on the prediction target $y$, for example in genomics \citep{marttinen2014assessing}. In general, PVE should be in $[0,1]$ because the predictions' variance should not exceed that of the data. However, this may not hold at test time for non-linear models such as neural networks if the models have overfitted to the training data, in which case the variance of the residual can exceed the variance of target in the test set. By placing a prior over $\mathbf{w}$ whose $\text{PVE}(\mathbf{w})$ concentrates around the PVE of the data generating process, the hypothesis space of the prior can be made more concentrated around the true model, which eventually yields a more accurate posterior. 

\subsection{Bayesian neural networks}
\subsubsection{Variational posterior approximation}
\label{sec: BNN}
Bayesian neural networks (BNNs) \citep{mackay1992practical,neal2012bayesian} are defined by placing a prior distribution on the weights $p(\mathbf{w})$ of a NN. Then, instead of finding point estimators of weights by minimizing a cost function, which is the normal practice in NNs, a posterior distribution of the weights is calculated conditionally on the data. Let $f(\mathbf{x};\mathbf{w})$ denote the output of a BNN and $p(y|\mathbf{x},\mathbf{w})=p(y|f(\mathbf{x};\mathbf{w}))$ the likelihood. Then, given a dataset of inputs $\mathbf{X}=\{\mathbf{x}^{(1)},\ldots,\mathbf{x}^{(N)}\}$ and outputs $\mathbf{y}=\{y^{(1)},\ldots,y^{(N)}\}$, training a BNN means computing the posterior distribution $p(\mathbf{w}|\mathbf{X},\mathbf{y})$. Variational inference can be used to approximate the intractable $p(\mathbf{w}|\mathbf{X},\mathbf{y})$ with a simpler distribution, $q_{\phi}(\mathbf{w})$, by minimizing $\text{KL}(q_{\phi}(\mathbf{w})||p(\mathbf{w}|\mathbf{X},\mathbf{y}))$. This is equivalent to maximizing the Evidence Lower BOund (ELBO) \citep{bishop2006pattern}
\begin{equation}
\begin{split}
    \mathcal{L}(\phi)=\mathcal{H}(q_{\phi}(\mathbf{w})) + \mathrm{E}_{q_{\phi}(\mathbf{w})}[\log p(\mathbf{y},\mathbf{w}|\mathbf{X})].
    \label{eq: elbo}
\end{split}
\end{equation}
The first term in Equation \ref{eq: elbo} is the entropy of the approximated posterior, which can be calculated analytically for many choices of $q_{\phi}(\mathbf{w})$. The second term is often estimated by the reparametrization trick \citep{kingma2013auto}, which reparametrizes the approximated posterior $q_{\phi}(\mathbf{w})$ using a deterministic and differentiable transformation $\mathbf{w}=g(\mathbf{\xi};\phi)$ with $\mathbf{\xi}\sim p(\xi)$, such that
$\mathrm{E}_{q_{\phi}(\mathbf{w})}[\log p(\mathbf{y},\mathbf{w}|\mathbf{X})]=\mathrm{E}_{p(\xi)}[\log p(\mathbf{y},g(\mathbf{\xi};\phi)|\mathbf{X})]$, which can be estimated by Monte Carlo integration.

\subsubsection{Gaussian scale mixture priors over weights}
The \textit{Gaussian scale mixture} (GSM) \citep{andrews1974scale} is defined to be a zero mean Gaussian conditional on its scales.
In BNNs, it has been combined with \textit{Automatic Relevance Determination} (ARD) \citep{mackay1996bayesian}, a widely used approach for feature selection in non-linear models. An ARD prior in BNNs means that all of the outgoing weights $w_{i,j}^{(l)}$ from node $i$ in layer $l$ share a same scale $\lambda_i^{(l)}$  \citep{neal2012bayesian}.
We define the input layer as layer $0$ for simplicity. A GSM ARD prior on each weight $w_{i,j}^{(l)}$ can be written in a \textit{hierarchically parametrized} form as follows:
\begin{equation}
w_{i,j}^{(l)}|\lambda_i^{(l)}, \sigma^{(l)}\sim \mathcal{N}(0, \sigma^{(l)2}\lambda_{i}^{(l)2});\; \;\lambda_{i}^{(l)}\sim p(\lambda_{i}^{(l)}; \theta_\lambda),
\label{eq: hierarchical parametrization}
\end{equation}
where $\sigma^{(l)}$ is the layer-wise global scale shared by all weights in layer $l$, which can either be set to a constant value or estimated using non-informative priors, and $p(\lambda_{i}^{(l)};\theta_\lambda)$ defines a hyper-prior on the local scales. The marginal distribution of $w_{i,j}^{(l)}$ can be obtained by integrating out the local scales given $\sigma^{(l)}$:
\begin{equation}
p(w_{i,j}^{(l)}|\sigma^{(l)})=\int \mathcal{N}(0, \sigma^{(l)2}\lambda_{i}^{(l)2})p(\lambda_{i}^{(l)};\theta_\lambda) \text{d}\lambda_{i}^{(l)}.
\label{eq: marginal distribution of weights}
\end{equation}
The hyper-prior of local scales $p(\lambda_{i}^{(l)};\theta_\lambda)$ determines the distribution of $p(w_{i,j}^{(l)}|\sigma^{(l)})$. For example, a Dirac delta distribution $\delta(\lambda_{i}^{(l)}-1)$ reduces $p(w_{i,j}^{(l)}|\sigma^{(l)})$ to a Gaussian with mean zero and variance $\sigma^{(l)2}$, whereas an inverse Gamma distribution on $\lambda_{i}^{(l)}$ makes $p(w_{i,j}^{(l)}|\sigma^{(l)})$ equal to the student's t-distribution \citep{gelman2013bayesian}. 

Many sparsity inducing priors in the Bayesian paradigm can be interpreted as Gaussian scale mixture priors with additional local scale variables $\tau_i^{(l)}$:
\begin{equation}
w_{i,j}^{(l)}|\lambda_i^{(l)}, \tau_i^{(l)}, \sigma^{(l)}\sim \mathcal{N}(0, \sigma^{(l)2}\lambda_{i}^{(l)2}\tau_{i}^{(l)2});\; \;\lambda_{i}^{(l)}\sim p(\lambda_{i}^{(l)}; \theta_\lambda);\; \;\tau_{i}^{(l)}\sim p(\tau_{i}^{(l)}; \theta_\tau).
\label{eq: general hierarchical parametrization}
\end{equation}
For example, the spike-and-slab prior \citep{mitchell1988bayesian} is the `gold standard' for sparse models and it introduces binary local scales $\tau_i^{(l)}$, interpreted as feature inclusion indicators, such that
\begin{equation}
w_{i,j}^{(l)}|\lambda_i^{(l)},\tau_i^{(l)},\sigma^{(l)}\sim (1 - \tau_i^{(l)})\delta(w_{i,j}^{(l)}) + \tau_i^{(l)}\mathcal{N}(0, \sigma^{(l)2}\lambda_{i}^{(l)2}),
\label{eq: spike-and-slab}
\end{equation}
where $\tau_i^{(l)}\sim \text{Bernoulli}(p)$. In Equation \ref{eq: general hierarchical parametrization}, the weight $w_{i,j}^{(l)}$ equals $0$ with probability $1-p$ (the spike) and with probability $p$ it is drawn from another Gaussian (the slab). Continuous local scales $\tau_i^{(l)}$ lead to other shrinkage priors, such as the horseshoe \citep{piironen2017hyperprior} and the Dirichlet-Laplace \citep{bhattacharya2015dirichlet}, which are represented as global-local (GL) mixtures of Gaussians.

\begin{figure}[t]
    \centering
    \includegraphics[width=.7\linewidth]{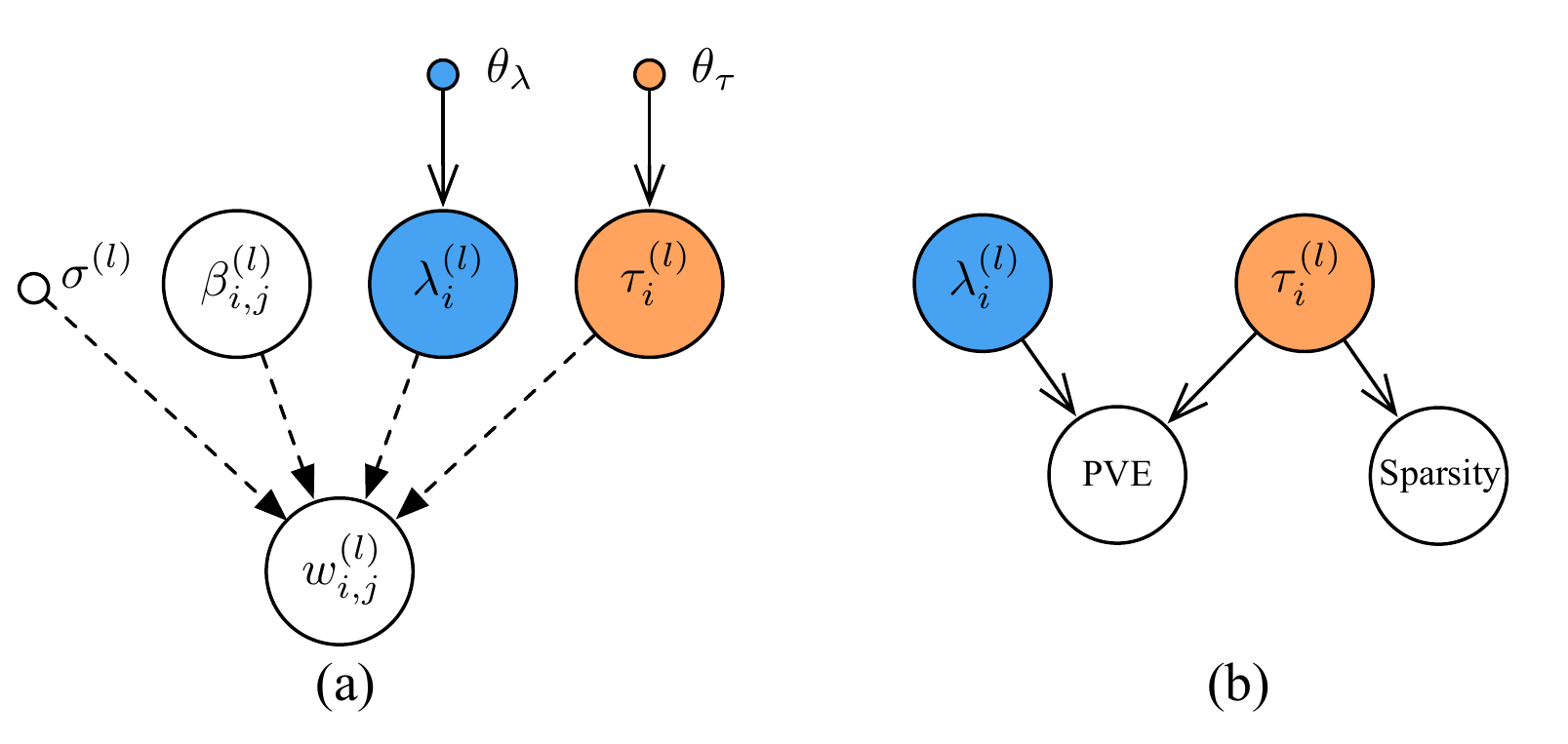}
    \caption{\textbf{a)} Non-centered parametrization of the GSM prior. \textbf{b)} The model's PVE is determined by $p(\lambda_{i}^{(l)};\theta_\lambda)$ and $p(\tau_{i}^{(l)};\theta_\tau)$ jointly, but sparsity is determined by $p(\tau_{i}^{(l)};\theta_\tau)$ alone. Therefore, we determine the distribution $p(\tau_{i}^{(l)};\theta_\tau)$ according to the prior knowledge about sparsity, and then tune $p(\lambda_{i}^{(l)};\theta_\lambda)$ conditionally on the previously selected $p(\tau_{i}^{(l)};\theta_\tau)$ to accommodate the prior knowledge about the PVE. }
    \label{fig: parametrization}
\end{figure}

Gaussian scale mixtures, i.e., Equation \ref{eq: general hierarchical parametrization}, are often written with an equivalent \textit{non-centered parametrization} \citep{papaspiliopoulos2007general} (Figure \ref{fig: parametrization}a),
\begin{equation}
w_{i,j}^{(l)}=\sigma^{(l)}\beta_{i,j}^{(l)}\lambda_{i}^{(l)}\tau_{i}^{(l)}; \; \beta_{i,j}^{(l)}\sim \mathcal{N}(0, 1);\; \lambda_{i}^{(l)}\sim p(\lambda_{i}^{(l)};\theta_\lambda);\; \tau_{i}^{(l)}\sim p(\tau_{i}^{(l)};\theta_\tau),
\label{eq: expanded parametrization}
\end{equation}
which has a better posterior geometry for inference \citep{betancourt2015hamiltonian} than the \textit{hierarchical parametrization}. Therefore, non-centered parametrization has been widely used in the BNN literature \citep{louizos2017bayesian,ghosh2018structured}, and we follow this common practice as well.

The hyper-parameter $\theta_\tau$ in $p(\tau_{i}^{(l)};\theta_\tau)$ controls the prior sparsity level, often quantified by the number of relevant features. However, for continuous hyper-priors, e.g. the half-Cauchy prior in the horseshoe, which do not force weights exactly to zero, it is not straightforward to select the hyper-parameter $\theta_\tau$ according to prior knowledge about sparsity. On the other hand, the existing discrete hyper-priors on $\tau_{i}^{(l)}$ model only restricted forms of sparsity, such as the Binomial distribution in the spike-and-slab prior in Equation \ref{eq: spike-and-slab}. In Section \ref{sec: sparsity info}, we propose an informative spike-and-slab prior consisting of a new class of discrete hyper-priors over the local scales $\tau_{i}^{(l)}$, capable of representing any type of sparsity. Moreover, the informative spike-and-slab makes $\tau_{i}^{(l)}$ dependent, which leads to a heavier penalization on false features than in the independent priors, such as the vanilla spike-and-slab, after correct features have been included.

It is well known that the scale parameter of the fully factorized Gaussian prior on BNNs weights affects the variability of the functions drawn from the prior \citep{neal2012bayesian}, and thus the PVE. When the PVE of the BNN has much probability around the correct PVE, the model is more likely to recover the true data generating mechanism (demonstration in Figure \ref{fig: illustration_pve}). As we will show in Section \ref{sec: PVE}, for a BNN with the GSM prior defined in Equation \ref{eq: expanded parametrization}, the hyper-priors on the local scales, $p(\lambda_{i}^{(l)};\theta_\lambda)$ and $p(\tau_{i}^{(l)};\theta_\tau)$, control the PVE jointly\footnote{The scale $\sigma^{(l)}$ is often estimated using a non-informative prior or cross-validated.} (Figure \ref{fig: parametrization}b). However, Figure \ref{fig: parametrization}b also shows how sparsity is determined by $p(\tau_{i}^{(l)};\theta_\tau)$ alone. Consequently, we propose choosing $p(\tau_{i}^{(l)};\theta_\tau)$ based on the prior knowledge on sparsity, and after that tuning the $p(\lambda_{i}^{(l)};\theta_\lambda)$ to achieve the desired level of the PVE, such that in the end our joint prior incorporates both types of prior knowledge.

\subsection{Stein Gradient Estimator}
\label{sec: SGE}
Stein Gradient Estimator (SGE) \citep{li2018gradient} provides a computational way to approximate the gradient of the log density (i.e., $\nabla_{\mathbf{z}}\log q(\mathbf{z})$), which only requires samples from $q(\mathbf{z})$ instead of its analytical form. Central for the derivation of the SGE is the Stein's identity \citep{liu2016kernelized}:
\begin{theorem}[Stein's identity]
\label{theorem: Stein's identity}
  Assume that $q(\mathbf{z})$ is a continuous differentiable probability density supported on $\mathcal{Z}\subset \mathbb{R}^{d}$, $\mathbf{h}:\mathcal{Z}\rightarrow\mathbb{R}^{d'}$ is a smooth vector-valued function $\mathbf{h}(\mathbf{z}) = [h_1(\mathbf{z}),\ldots,h_{d'}(\mathbf{z})]^T$, and $\mathbf{h}$ is in the Stein class of $q$ such that
\begin{equation}
\lim_{\mathbf{z}\rightarrow\infty}q(\mathbf{z})\mathbf{h}(\mathbf{z})=0 \;\; \text{if} \;\; \mathcal{Z} = \mathbb{R}^{d}.
\label{eq: stein class}
\end{equation}
Then the following identity holds:
\begin{equation}
\mathbb{E}_{q}[\mathbf{h}(\mathbf{z})\nabla_{\mathbf{z}}\log q(\mathbf{z})^{T}+\nabla_{\mathbf{z}}\mathbf{h}(\mathbf{z})]=0.
\label{eq: stein's identity}
\end{equation}
\end{theorem}
SGE estimates $\nabla_{\mathbf{z}}\log q(\mathbf{z})$ by inverting Equation \ref{eq: stein's identity} and approximating the expectation with $K$ Monte Carlo samples $\{\mathbf{z}^{(1)},\ldots,\mathbf{z}^{(K)}\}$ from $q(\mathbf{z})$, such that
\begin{equation}
-\frac{1}{K}\mathbf{HG}\approx\overline{\nabla_{\mathbf{z}}\mathbf{h}},
\label{eq: approximate stein's identity}
\end{equation}
where $\mathbf{H} = (\mathbf{h}(\mathbf{z}^{(1)}),\ldots,\mathbf{h}(\mathbf{z}^{(K)}))\in\mathbb{R}^{d'\times K}$, $\overline{\nabla_{\mathbf{z}}\mathbf{h}}=\frac{1}{K}\sum_{k=1}^{K}\nabla_{\mathbf{z}^{(k)}}\mathbf{h}(\mathbf{z}^{(k)})\in \mathbb{R}^{d'\times d}$, and the matrix $\mathbf{G}=(\nabla_{\mathbf{z}^{(1)}}\log q(\mathbf{z}^{(1)}),\ldots,\nabla_{\mathbf{z}^{(K)}}\log q(\mathbf{z}^{(K)}))^{T}\in \mathbb{R}^{K\times d}$ contains the gradients of $\nabla_{\mathbf{z}}\log q(\mathbf{z})$ for the $K$ samples. Thus a ridge regression estimator is designed to estimate $G$ by adding an $l_2$ regularizer:
\begin{equation}
\hat{\mathbf{G}}^{\text{Stein}}=\argmin_{\mathbf{G}\in \mathbb{R}^{K\times d}}||\overline{\nabla_{\mathbf{z}}\mathbf{h}} + \frac{1}{K}\mathbf{HG}||^{2}_{F}+\frac{\eta}{K^2}||\mathbf{G}||^{2}_{F},
\label{eq: loss function of SGE}
\end{equation}
where $||\cdot||_{F}$ is the Frobenius norm of a matrix and the penalty $\eta\geq0$. By solving Equation \ref{eq: loss function of SGE}, the SGE is obtained:
\begin{equation}
\hat{\mathbf{G}}^{\text{Stein}}=-K(\mathbf{K}+\eta\mathbf{I})^{-1}\mathbf{H}^{T}\overline{\nabla_{\mathbf{z}}\mathbf{h}},
\label{eq: SGE estimator}
\end{equation}
where $\mathbf{K} = \mathbf{H}^{T}\mathbf{H}$ is the kernel matrix, such that $\mathbf{K}_{i,j}=\mathcal{K}(\mathbf{z}^{(i)}, \mathbf{z}^{(j)}) = \mathbf{h}(\mathbf{z}^{(i)})^T\mathbf{h}(\mathbf{z}^{(j)})$, and $(\mathbf{H}^{T}\overline{\nabla_{\mathbf{z}}\mathbf{h}})_{i,j} = \sum_{k=1}^{K}\nabla_{z_{j}^{(k)}}\mathcal{K}(\mathbf{z}^{(i)}, \mathbf{z}^{(k)})$, where $\mathcal{K}(\cdot, \cdot)$ is the kernel function. It has been shown that the RBF kernel satifies Stein's identity and is a default choice for Stein Gradient Estimator. 

\section{Prior knowledge about sparsity}
\label{sec: sparsity info}
In this section, we propose a new hyper-prior for the local scales $p(\tau_i^{(l)};\theta_\tau)$ to model prior beliefs about sparsity. The new prior generates the local scales conditionally on the number of relevant features, which allows us to explicitly express prior knowledge about the number of relevant features. We focus on the case where each local scale $\tau_i^{(l)}$ is assumed to be binary with domain $\{0,1\}$, analogously to the feature inclusion indicators in the spike-and-slab prior.

\subsection{Prior on the number of relevant features}
\begin{figure*}[ht]
    \centering
    \includegraphics[width=.5\linewidth]{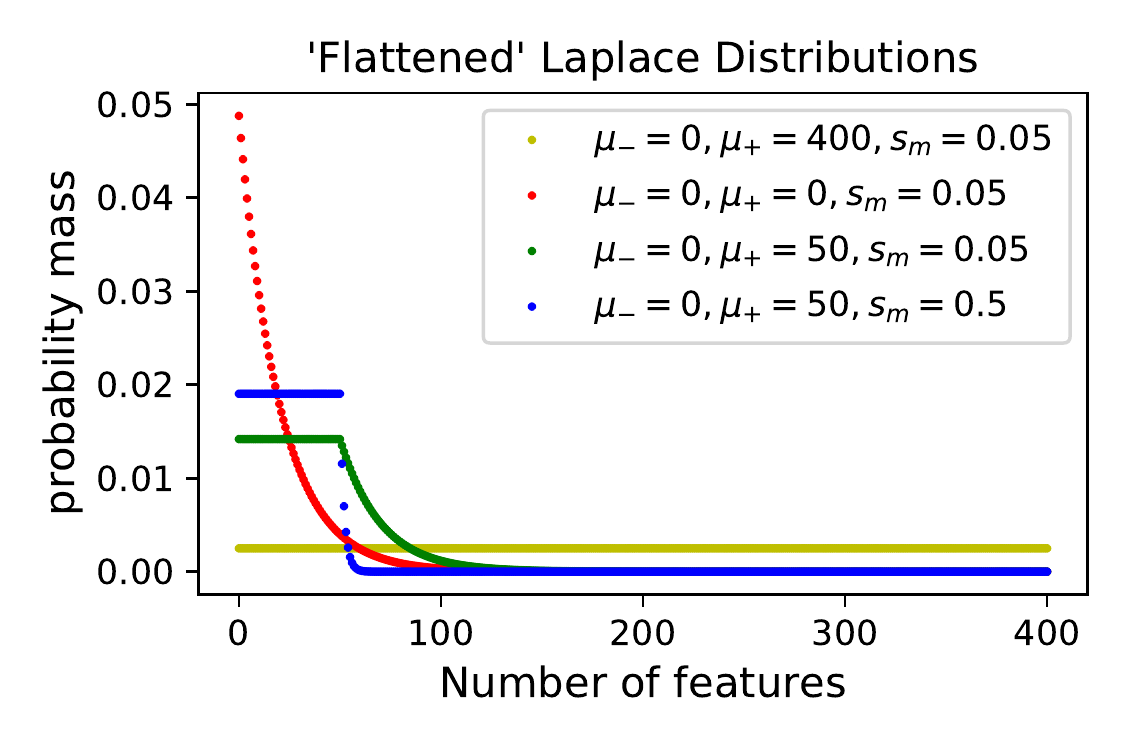}
    \caption{A visualization of 'flattened' Laplace prior with different hyper-parameters. 'Discretized' exponential (red) and uniform (yellow) distributions are special cases of the 'flattened' Laplace. 
    }
    \label{fig: visualization of FL}
\end{figure*}

We control sparsity by placing a prior on the number of relevant features $m$ using a probability mass function $p_m(m;\theta_m)$, where $0\leq m\leq D$ (dimension of the dataset). Intuitively, if $p_m$ concentrates close to $0$, a sparse model with few features is preferred; if $p_m$ places much probability mass close to $D$, then all of the features are likely to be used instead. Hence, unlike other priors encouraging shrinkage, such as the horseshoe, our new prior easily incorporates experts' knowledge about the number of relevant features.
In practice, $p_m(m;\theta_m)$ is chosen based on the available prior knowledge. When there is a good idea about the number of relevant features, a unimodal distribution, such as a discretized Laplace, can be used:
\begin{equation}
p_m(m;\mu, s_m)=c_n\exp\{-\frac{s_m|m-\mu_m|}{2}\},
\label{eq: discrete laplace}
\end{equation}
where $\mu_m$ is the mode, $s_m$ is the precision, and $c_n$ is the normalization constant. Often only an interval for the number of relevant features is available. Then it is possible to use, for example, a `flattened' Laplace (Figure \ref{fig: illustration_sparsity}):
\begin{equation}
\begin{split}
&p_m(m;\mu_{-},\mu_{+},  s_m)=c_n\exp\{-\frac{s_m\mathcal{R}(m; \mu_{+}, \mu_{-})}{2}\},\\
&\mathcal{R}(m;  \mu_{-}, \mu_{+}) = \max\{(m-\mu_{+}), (\mu_{-}-m), 0\},
\label{eq: flattened laplace}
\end{split}
\end{equation}
where $[\mu_{-}, \mu_{+}]$ defines the interval where the probability is uniform and reaches its maximum value, and $c_n$ is the corresponding normalization constant. Equation \ref{eq: discrete laplace} and \ref{eq: flattened laplace} include the (discretized) exponential distribution as a special case with $\mu_m=0$ and $\mu_{-}=\mu_{+}=0$ respectively; it has been widely studied in sparse deep learning literature \citep{polson2018posterior,wang2020uncertainty}. The `flattened' Laplace, with a high precision $s_m$, is a continuous approximation of the distribution with a uniform probability mass within $[\mu_{-},\mu_{+}]$ and 0 outside of $[\mu_{-},\mu_{+}]$ (blue in  Figure \ref{fig: visualization of FL}). If there is no prior information about sparsity, a discrete uniform prior over $[0,D]$ is a plausible alternative. See Figure \ref{fig: visualization of FL} for a visualization. 

\subsection{Feature allocation}
Conditionally on the number of features $m$, we introduce indicators $I_i\in\{0,1\}$ to denote if a feature $i$ is used by the model ($I_i = 1$) or not ($I_i = 0$), such that $m = \sum_{i=1}^{D}I_{i}$. We then marginalize over $m$ using $p_m(m;\theta_m)$. We assume there is no prior knowledge about relative importance of features (this assumption can be easily relaxed if needed), i.e., $\{I_i\}_{i=1}^{D}$ has a jointly uniform distribution given $m$:
\begin{equation}
p(\{I_i\}_{i=1}^{D}|m)=c_d\delta[m-\sum_{i=1}^{D}I_{i}],
\label{eq: conditional uniform discrete}
\end{equation}
where the normalization constant is $c_d = {D \choose m}^{-1}$.
Now we can calculate the joint distribution of $\{I_i\}_{i=1}^{D}$ by marginalizing out $m$ :
\begin{equation}
\begin{split}
p(\{I_i\}_{i=1}^{D};\theta_m) &= \sum_{m=0}^{D} p_m(m;\theta_m)p(\{I_i\}_{i=1}^{D}|m)=p_m(\sum_{i=1}^{D}I_{i};\theta_m){D \choose \sum_{i=1}^{D}I_{i}}^{-1}.
\label{eq: joint indentity}
\end{split}
\end{equation}
When the local scale variables $\tau_i^{(l)}$ are binary, the $\tau_i^{(l)}$ take the role of the identity variables $I_i$. Thus we obtain a joint distribution over discrete scale parameters $\tau_{i}$ as
\begin{equation}
p(\tau_1^{(l)},\ldots,\tau_D^{(l)};\theta_\tau)=p_m(\sum_{i=1}^{D}\tau_i^{(l)};\theta_m){D \choose \sum_{i=1}^{D}\tau_i^{(l)}}^{-1},
\label{eq: joint scale}
\end{equation}
where $\theta_\tau$ represents the same set of hyper-parameters as $\theta_m$, and the distribution $p_m(\sum_{i=1}^{D}\tau_i^{(l)};\theta_m)$ models the beliefs of the number of relevant features.
In general, the local scales $\{\tau_i^{(l)}\}_{i=1}^{D}$ in Equation \ref{eq: joint scale} are dependent. However, when $p_m(\cdot)$ is set to a Binomial (or its Gaussian approximation), the joint distribution of $\{\tau_i^{(l)}\}_{i=1}^{D}$ factorizes into a product of independent Bernoullis corresponding to the vanilla spike-and-slab with a fixed slab probability (Equation \ref{eq: spike-and-slab}). We refer to Equation \ref{eq: joint scale} as the \textit{informative spike-and-slab} prior.

In BNNs, we suggest to use the informative spike-and-slab prior on the first layer to inform the model about sparsity on the feature level. For hidden layers, we do not assume prior domain knowledge, as they encode latent representations where such knowledge is rare. However, a uniform prior on the number of hidden nodes can be applied on hidden layers to infer optimal layer sizes from data by computing the posteriors, which we leave for future work. In this work, for the hidden layers we use the standard Gaussian scale mixture priors in Equation \ref{eq: hierarchical parametrization}. 

\section{Prior knowledge on the PVE}
\label{sec: PVE}
After incorporating prior knowledge about sparsity in the new informative hyper-prior $p(\tau_i^{(l)};\theta_\tau)$, in this section we introduce an optimization approach to determine the hyper-parameters (i.e., $\theta_\lambda$) of the hyper-prior for the other local scale parameters $p(\lambda_i^{(l)};\theta_\lambda)$ in the GSM prior (Equation \ref{eq: expanded parametrization}), based on domain knowledge about the PVE.

\subsection{PVE for Bayesian neural networks}
According to the definition of PVE, when $f(\mathbf{x};\mathbf{w})$ is a BNN, $\text{PVE}(\mathbf{w})$ has a distribution induced by the distribution on $\mathbf{w}$ via Equation \ref{eq: PVE}. We denote the variance of the unexplainable noise $\epsilon$ in the regression model in Equation \ref{eq: data generating model} by $\sigma_\epsilon^2$.
We use $\mathbf{w}_{(\boldsymbol{\sigma}, \theta_\lambda, \theta_\tau)}$ to denote the BNN weights with a GSM prior (i.e., Equation \ref{eq: expanded parametrization}) parametrized by hyper-parameters $\{\boldsymbol{\sigma}, \theta_\lambda, \theta_\tau\}$, where $\boldsymbol{\sigma}$ is $\{\sigma^{(0)},\ldots,\sigma^{(L)}\}$.
The PVE of a BNN with a GSM prior is given by
\begin{equation}
\text{PVE}(\mathbf{w}_{(\boldsymbol{\sigma}, \theta_\lambda, \theta_\tau)},\sigma_\epsilon) = \frac{\text{Var}(f(\mathbf{X};\mathbf{w}_{(\boldsymbol{\sigma}, \theta_\lambda, \theta_\tau)}))}{\text{Var}(f(\mathbf{X};\mathbf{w}_{(\boldsymbol{\sigma}, \theta_\lambda, \theta_\tau)}))+\sigma_\epsilon^2}.
\label{eq: PVE of BNN}
\end{equation}
The noise $\sigma_\epsilon$ and layer-wise global scales $\boldsymbol{\sigma}$ are usually given the same non-informative priors  \citep{zhang2018variable} or set to a default value \citep{blundell2015weight}. Setting of the hyper-parameter $\theta_\tau$ was described in Section 3.
Therefore, we then optimize the remaining hyper-parameter $\theta_\lambda$ to make the distribution of the PVE match our prior knowledge about the PVE.


\subsection{Optimizing hyper-parameters according to prior PVE}
Denote the available prior knowledge about the PVE by $p(\rho)$. In practice such a prior may be available from previous studies, and here we assume it can be encoded in to the reasonably flexible Beta distribution. If no prior information is available, a uniform prior, i.e., Beta$(1,1)$, can be used. We incorporate such knowledge into the prior by optimizing the hyper-parameter $\theta_\lambda$ such that the distribution induced by the BNN weight prior, $p(\mathbf{w}; \boldsymbol{\sigma}, \theta_\lambda, \theta_\tau)$, on the BNN model's PVE denoted by $q_{\theta_\lambda}(\rho(\mathbf{w}))$\footnote{Hyper-parameters $\sigma$ and $\theta_\tau$ are omitted for simplicity as they are not optimized.}, is close to $p(\rho)$. 
We achieve this by minimizing the Kullback–Leibler divergence from $q_{\theta_\lambda}(\rho(\mathbf{w}))$ to $p(\rho)$ w.r.t. the hyper-parameter $\theta_\lambda$, i.e., $\theta_\lambda^{*}=\argmin_{\theta_\lambda}\text{KL}[q_{\theta_\lambda}(\rho(\mathbf{w}))|p(\rho)]$. However, the KL divergence is not analytically tractable because the $q_{\theta_\lambda}(\rho(\mathbf{w}))$ is defined implicitly, such that we can only sample from $q_{\theta_\lambda}(\rho(\mathbf{w}))$ but can not evaluate its density. We first observe that the KL divergence can be approximated by:
\begin{equation}
\small
\begin{split}
\text{KL}[q_{\theta_\lambda}(\rho(\mathbf{w}))|p(\rho)] &= \text{E}_{p(\mathbf{w};\theta_\lambda)}\left[ \log\frac{q_{\theta_\lambda}(\rho(\mathbf{w}))}{p(\rho(\mathbf{w}))} \right] = \text{E}_{p(\xi)}\left[\log\frac{q(\rho(g(\xi; \theta_\lambda)))}{p(\rho(g(\xi; \theta_\lambda)))}\right]\\
&\approx
\frac{1}{M}\sum_{m=1}^{M}\log q_{\theta_\lambda}(\rho(g(\xi^{(m)}; \theta_\lambda))) - \log p(\rho(g(\xi^{(m)}; \theta_\lambda))),
\label{eq: MC estimation of KL}
\end{split}
\end{equation}
by reparametrization and Monte Carlo integration. Here we assume that the GSM distribution $p(\mathbf{w}; \theta_\lambda)$ can be reparametrized by a deterministic function $\mathbf{w}=g(\mathbf{\xi}; \theta_\lambda)$ with $\mathbf{\xi}\sim p(\xi)$. For non-reparametrizable distributions, score function estimators can be used instead, which we leave for future work. Moreover, since $\text{PVE}(\mathbf{w})$ is another deterministic function of $\mathbf{w}$ given data $\mathbf{X}$, we have $\text{PVE}(\mathbf{w})= \rho(\mathbf{w}) = \rho(g(\mathbf{\xi}; \theta_\lambda))$. The expectation is approximated by $M$ samples from $p(\xi)$. Then the gradient of the KL w.r.t. $\theta_\lambda$ can be calculated by:
\begin{equation}
\small
\begin{split}
\nabla_{\theta_\lambda}\text{KL}\left[q_{\theta_\lambda}(\rho(\mathbf{w}))|p(\rho)\right] &\approx
\frac{1}{M}\sum_{m=1}^{M}\nabla_{\theta_\lambda}\left[\log q_{\theta_\lambda}(\rho(g(\xi^{(m)}; \theta_\lambda))) - \log p(\rho(g(\xi^{(m)}; \theta_\lambda)))\right]\\
&=\frac{1}{M}\sum_{m=1}^{M}\nabla_{\theta_\lambda}\rho(g(\xi^{(m)}; \theta_\lambda))\left[\nabla_{\rho}\log q_{\theta_\lambda}(\rho) - \nabla_{\rho}\log p(\rho)\right].
\label{eq: gradient of KL}
\end{split}
\end{equation}
The first term $\nabla_{\theta_\lambda}\rho(g(\xi^{(m)}; \theta_\lambda))$ can be calculated exactly with back-propagation packages, such as PyTorch. The last term, the gradient of the log density $\nabla_{\rho}\log p(\rho)$, is tractable for a prior with a tractable density, such as the Beta distribution. However, the derivative $\nabla_{\rho}\log q_{\theta_\lambda}(\rho)$ is generally intractable, as the distribution of the PVE of a BNN $q_{\theta_\lambda}(\rho)$ is implicitly defined by Equation \ref{eq: PVE of BNN}. We propose to apply SGE (Section \ref{sec: SGE}) to estimate $\nabla_{\rho}\log q_{\theta_\lambda}(\rho)$, which only requires samples from $q_{\theta_\lambda}(\rho)$. 

When noise $\sigma_{\epsilon}$ and layer-wise global scales $\boldsymbol{\sigma}$ are given non-informative priors, e.g., $\text{Inv-Gamma}(0.001,0.001)$, drawing samples directly according to Equation \ref{eq: PVE of BNN} is unstable, because the variance of the non-informative prior does not exist. Fortunately, if all activation functions of the BNN are positively homogeneous (e.g., ReLU), we have the following theorem (proof is given in Supplementary):
\begin{theorem}
\label{theorem: Variance of BNN with GSM prior}
Assume an L-layer BNN with Gaussian scale mixture prior in the form of Equation \ref{eq: expanded parametrization}, if all activation functions are positively homogeneous, e.g., ReLU, we have:
\begin{equation}
\text{Var}(f(\mathbf{X};\mathbf{w}_{(\boldsymbol{\sigma}, \theta_\lambda, \theta_\tau)})) = \text{Var}(f(\mathbf{X};\mathbf{w}_{(1, \theta_\lambda, \theta_\tau)}))\prod_{l=0}^{L}\sigma^{(l)2}.
\label{eq: variance BNN}
\end{equation}
\end{theorem}
\noindent Now instead of giving non-informative priors to all layer-wise global scales, i.e., $\sigma_{\epsilon}=\sigma^{(0)}=\ldots=\sigma^{(L)}\sim \text{Inv-Gamma}(0.001,0.001)$, we propose to use $\sigma^{(0)}=\ldots=\sigma^{(L-1)}=1$, and $\sigma_{\epsilon}=\sigma^{(L)}\sim \text{Inv-Gamma}(0.001,0.001)$.
By substituting Equation \ref{eq: variance BNN} with these specifications into Equation \ref{eq: PVE of BNN}, we can write the PVE as
\begin{equation}
\text{PVE}(\mathbf{w}_{(\sigma, \theta_\lambda, \theta_\tau)},\sigma_\epsilon) = \frac{\text{Var}(f(\mathbf{X};\mathbf{w}_{(1, \theta_\lambda, \theta_\tau)}))}{\text{Var}(f(\mathbf{X};\mathbf{w}_{(1, \theta_\lambda, \theta_\tau)}))+\frac{\sigma_{\epsilon}^2}{\sigma^{(L)2}}} = \frac{\text{Var}(f(\mathbf{X};\mathbf{w}_{(1, \theta_\lambda, \theta_\tau)}))}{\text{Var}(f(\mathbf{X};\mathbf{w}_{(1, \theta_\lambda, \theta_\tau)}))+1},
\label{eq: simple PVE of BNN}
\end{equation}
which avoids sampling from the non-informative inverse Gamma distribution.
\begin{figure*}[t]
    \centering
    \includegraphics[width=1.\linewidth]{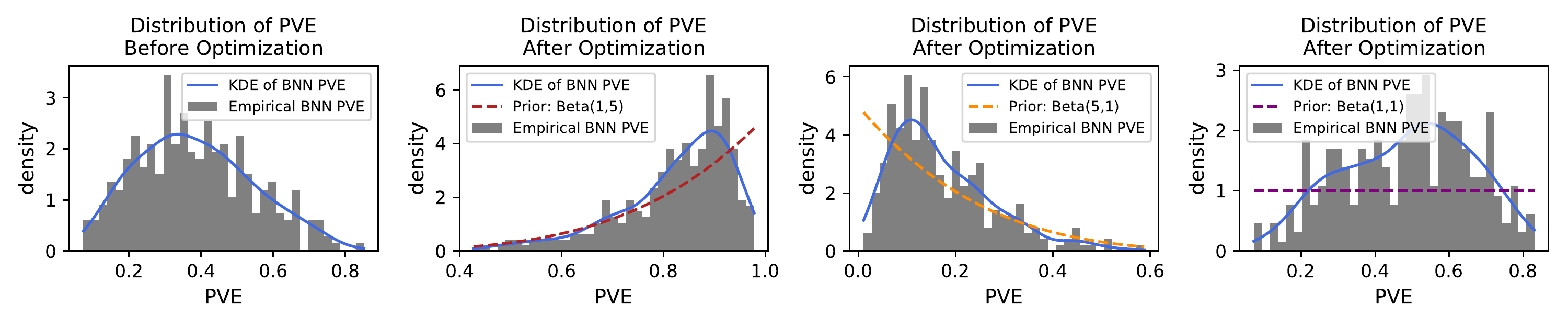}
    \includegraphics[width=1.\linewidth]{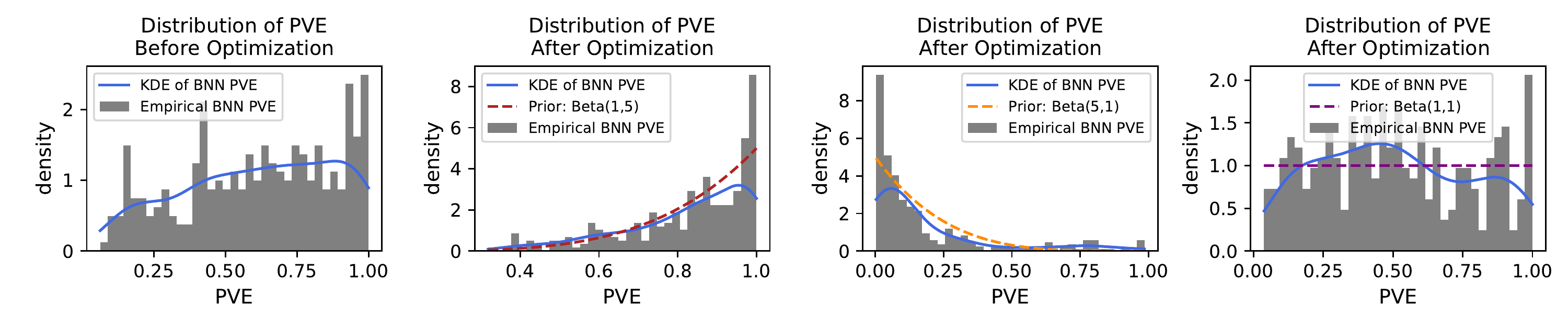}
    \caption{Empirical distributions (50 samples) of the BNN's PVE before (first column) and after optimizing hyper-parameters according to three different prior PVEs: Beta(1,5), Beta(5,1) and Beta(1,1), shown in the last three columns. Top: Results for mean-field Gaussian prior, where the local scale is optimized. Bottom: Results for hierarchical Gaussian prior, where the hyper-parameter of the inverse gamma prior is optimized.
    }
    \label{fig: PVE learning}
\end{figure*}

Figure \ref{fig: PVE learning} illustrates the proposed approach, where we applied the method on two 3-layer BNNs containing 100-50-30-1 nodes and the ReLU activation. We considered two GSM ARD priors for the BNN weights: a mean-field Gaussian prior
\begin{equation}
\begin{split}
w_{i,j}^{(l)}|\lambda_i^{(l)}\sim\mathcal{N}(0,\sigma^{(l)2}\lambda_{i}^{(l)2}), \lambda_{i}^{(l)}=\sigma_\lambda,
\end{split}
\end{equation}
and a hierarchical Gaussian prior
\begin{equation}
\begin{split}
w_{i,j}^{(l)}|\lambda_i^{(l)}\sim\mathcal{N}(0,\sigma^{(l)2}\lambda_{i}^{(l)2}), \lambda_{i}^{(l)}\sim \text{Inv-Gamma}(\alpha, \beta),
\end{split}
\end{equation}
with the non-informative prior on the noise and the last layer-wise global scale. For the Gaussian prior, we optimized $\sigma_\lambda$ according to the prior PVE. For the hierarchical Gaussian prior, we optimized $\beta$ while fixing $\alpha=2$ because the shape parameter of the Gamma distribution is non-reparametrizable. We see that after optimizing the hyperpriors, the simulated empirical distributions of the PVE is close to the corresponding prior knowledge in both cases, especially when the prior knowledge is informative, such as the Beta(1, 5) or Beta(5, 1). Moreover, we observe that even when we have fixed the shape parameter $\alpha$ of the inverse Gamma, the prior is still flexible enough to approximate the prior PVE well.

\section{Related literature}\label{sec:related}
\textbf{Priors on the number of relevant features} have been applied on small datasets to induce sparsity in shallow models, e.g. NNs with one hidden layer \citep{vehtari2001bayesian}, including the geometric \citep{insua1998feedforward} and  truncated Poisson \citep{denison1998automatic, insua1998feedforward, andrieu2000robust, kohn2001nonparametric} distributions. However, those approaches rely on the reversible jump Markov chain Monte Carlo (RJMCMC) to approximate the posterior \citep{phillips1996bayesian, sykacek2000input, andrieu2013reversible}, which does not scale up to deep architectures and large datasets. In this work, we incorporate such prior beliefs into the hyper-prior on the local scales of the Gaussian scale mixture prior; thus, the posterior can be approximated effectively by modern stochastic variational inference \citep{hoffman2013stochastic}, even for deep NN architectures and large datasets.

\noindent\textbf{Priors on PVE} have been proposed to fine-tune hyper-parameters \citep{zhang2018variable} and to construct shrinkage priors \citep{zhang2020bayesian} for Bayesian linear regression, where the distribution on the model PVE is analytically tractable. \cite{marttinen2014assessing} used simulation and grid search to incorporate prior knowledge about a point estimate for the PVE in Bayesian reduced rank regression, and used it to assess the impact of multiple genetic covariates on multivariate targets. No previous work has applied and evaluated incorporating domain knowledge on the PVE in NN priors, where the distribution of the model's PVE is defined implicitly.

\noindent\textbf{Priors on the BNNs weights:} BNNs with a fully factorized Gaussian prior were proposed by \cite{graves2011practical} and \cite{blundell2015weight}. They can be interpreted as NNs with dropout by using a mixture of Dirac-delta distributions to approximate the posterior \citep{gal2016dropout}. \cite{nalisnick2019dropout} extended these works and showed that NNs with any multiplicative noise could be interpreted as BNNs with GSM ARD priors. Priors over weights with low-rank assumptions, such as the k-tied normal \citep{swiatkowski2020k} and rank-1 perturbation \citep{dusenberry2020efficient} were found to have better convergence rates and ability to capture multiple modes when combined with ensemble methods. Matrix-variate Gaussian priors were proposed by \cite{neklyudov2017structured} and \cite{sun2017learning} to improve the expressiveness of the prior by accounting for the correlations between the weights. Some priors have been proposed to induce sparsity, such as the log-uniform \citep{molchanov2017variational, louizos2017bayesian}, log-normal \citep{neklyudov2017structured}, horseshoe \citep{louizos2017bayesian, ghosh2018structured}, and spike-and-slab priors \citep{deng2019adaptive}. However, none of the works has proposed how to explicitly encode domain knowledge into the prior on the NN weights.
  
\noindent\textbf{Informative priors of BNNs:} Building of informative priors for NNs has been studied in the function space. One common type of prior information concerns the behavior of the output with certain inputs. Noise contrastive priors (NCPs) \citep{hafner2018noise} were designed to encourage reliable high uncertainty for OOD (out-of distribution) data points. Gaussian processes were proposed as a way of defining functional priors because of their ability to encode rich functional structures. \cite{flam2017mapping} transformed a functional GP prior into a weight-space BNN prior, with which variational inference was performed. Functional BNNs \citep{sun2019functional} perform variational inference directly in the functional space, where meaningful functional GP priors can be specified. \cite{pearce2019expressive} used a combination of different BNN architectures to encode prior knowledge about the function. Although functional priors avoid working with uninterpretable high-dimensional weights, encoding sparsity of features into the functional space is non-trivial.

\section{Experiments}\label{sec:experiments}
In this section, we first compare the proposed informative sparse prior with alternatives in a feature selection task on synthetic toy data sets. We then apply it on seven public UCI real-world datasets\footnote{https://archive.ics.uci.edu/ml/index.php}, where we tune the level of noise and the fraction of informative features. Finally, in a genetics application we show that incorporating domain knowledge on both sparsity and the PVE can improve results in a realistic scenario.


\subsection{Synthetic data}
\textbf{Setup:} We first validate the performance of the informative spike-and-slab prior proposed in Section \ref{sec: sparsity info} on a feature selection task, using  synthetic datasets similar to those discussed by \cite{van2014horseshoe} and \cite{piironen2017sparsity}. Instead of a BNN, we here use linear regression, i.e., a NN without hidden layers, which enables comparing the proposed strategy of encouraging sparsity with existing alternatives.

Consider $n$ datapoints generated by:
\begin{equation}
y_i= w_i +\epsilon_i,\;\;\;\epsilon_i\sim\mathcal{N}(0,\sigma_\epsilon^2),\;\;\;i=1,\ldots,n,
\label{eq: simulator}
\end{equation}
where each observation $y_i$ is obtained by adding Gaussian noise with variance $\sigma_\epsilon^2$ to the signal $w_i$. We set the number of datapoints $n$ to $400$, the first $p_0=20$ signals $\{w_i|i=1,\ldots,20\}$ equal to $A$ (signal levels), and the rest of the signals to 0. We consider 3 different noise levels $\sigma_\epsilon\in\{1, 1.5, 2\}$, and 10 different signal levels $A\in\{1, 2,\ldots, 10\}$. For each parameter setting (30 in all), we generate 100 data realizations. The model in Equation \ref{eq: simulator} can be considered a linear regression: $\mathbf{y} = X\mathbf{\hat{w}}^{T}+\boldsymbol{\epsilon}$, where $X = I$ and $\mathbf{\hat{w}} = (A,\ldots,A,0,0,\ldots,0)$ with the first $p_0$ elements being $A$, so this is a feature selection task where the number of features and datapoints are equal. We use the mean squared error (MSE) between the posterior mean signal $\bar{\mathbf{w}}$ and the true signal $\mathbf{\hat{w}}$ to measure the performance.
\begin{table}[t]
\caption{Details of the four Gaussian scale mixture priors included in the comparison on the synthetic datasets. The vague Inv-Gamma represents a diffuse inverse gamma prior Inv-Gamma(0.001,0.001). FL denotes the `flattened' Laplace prior defined in Equation \ref{eq: flattened laplace}. NA means that the model is defined without the corresponding hyperprior.}
\label{tab: feature selection model}
\begin{tabular}{c|c|c|c}
\hline
Name                    & $p(\lambda_i)$ & $p(\tau_i)$\footnote{For the horseshoe, scale $\tau$ is shared by all weights.} & Hyper-prior \\ \hline
 \texttt{BetaSS}          & vague Inv-Gamma & Bernoulli($p$) &  $p\sim \text{Beta}(2, 38)$           \\ \hline
 \texttt{DeltaSS}     & vague Inv-Gamma & Bernoulli($p$) &  $p=0.05$           \\ \hline
 \texttt{InfoSS}   & vague Inv-Gamma & FL($0,30,5$) &   NA         \\ \hline
\texttt{HS}               & $C^{+}(0,1)$& $\tau\sim C^{+}(0,(\frac{p_0}{n-p_0})^2)$  &    NA         \\ \hline
\end{tabular}
\end{table}
\begin{figure*}[h!]
    \centering
    \includegraphics[width=1.\linewidth]{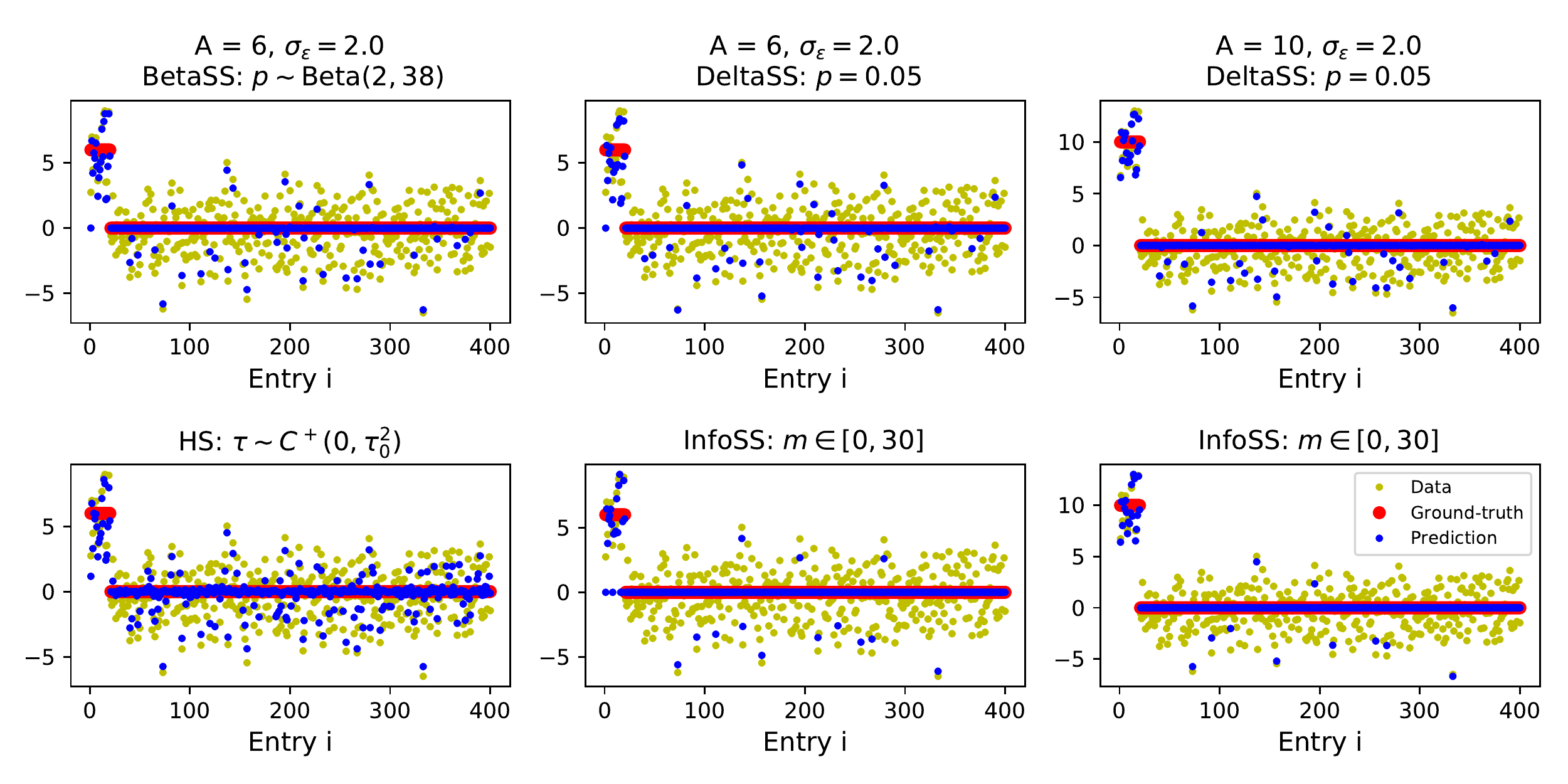}
    \caption{\textit{Synthetic datasets}: A detailed visualization of the features selected by the four priors with noise level $\sigma=2$ and signal level $A=6$ (first two columns).  
    Results for the two best models (\texttt{DeltaSS} and \texttt{InfoSS}) are additionally shown with a larger signal $A=10$ (the last column). Beta spike-and-slab (\texttt{BetaSS}) is slightly worse than Delta spike-and-slab (\texttt{DeltaSS}), because the latter uses the correct slab probability. Informative spike-and-slab (\texttt{InfoSS}) outperforms alternatives by making the signals dependent. Horseshoe (\texttt{HS}) with the correct sparsity level overfits to the noise.
    }
    \label{fig: feature selection visualization}
\end{figure*}

\noindent\textbf{Parameter settings:} For estimation we use linear regression with the correct structure.
We consider 4 different Gaussian scale mixture priors on $\mathbf{w}$ that all follow the general form
\begin{equation}
w_i\sim\mathcal{N}(0,\sigma^2\lambda_i^2\tau_i^2),\;\;\sigma=1,\;\;\lambda_i\sim p(\lambda_i),\;\;\tau_i\sim p(\tau_i).
\label{eq: GSM for linear regression}
\end{equation}
The details of the different priors are shown in Table \ref{tab: feature selection model}. For all the spike-and-slab (\texttt{SS}) priors, we place a diffuse inverse Gamma prior on $p(\lambda_i)$. For the \texttt{BetaSS} and \texttt{DeltaSS} priors, we assume that $p(\tau_i)=\text{Bernoulli}(p)$, and define $p\sim \text{Beta}(2,38)$ for \texttt{BetaSS} and $p=0.05$ for \texttt{DeltaSS}, which both reflect the correct level of sparsity. For the informative spike-and-slab, \texttt{InfoSS}, we use the 'flattened' Laplace (FL) prior defined in Equation \ref{eq: flattened laplace} with $\mu_{-}=0, \mu_{+}=30$, and $s_m=5$, to encode prior knowledge that the number of non-zero signals is (approximately) uniform on $[0, 30]$. We place an informative half-Cauchy prior $C^{+}(0,\tau_0^2)$ on the global shrinkage scale $\tau$ with $\tau_0 = \frac{p_0}{n-p_0}$ in the horseshoe (\texttt{HS}) \citep{piironen2017sparsity}, to assume the same sparsity level as the other priors\footnote{The effective number of features \citep{piironen2017sparsity} is set to $20$.}. 

\begin{figure*}[t]
    \centering
    \includegraphics[width=1.\linewidth]{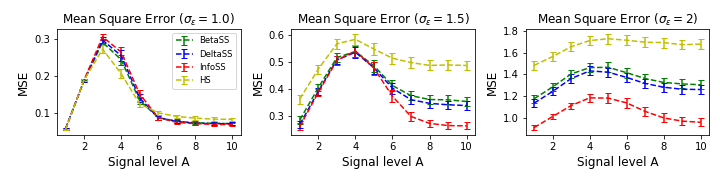}
    \caption{\textit{Synthetic datasets}: Mean squared error (MSE) between the estimated and true signals. The bars represent the $95\%$ confidence intervals over 100 datasets. The novel \texttt{InfoSS} prior is indistinguishable from the other SS priors when for the noise level is low. However,  \texttt{InfoSS} is significantly more accurate than the other priors when there is more noise.
    }
    \label{fig: feature selection MSE}
\end{figure*}

\noindent\textbf{Results:} Figures \ref{fig: feature selection visualization} and \ref{fig: feature selection MSE} show the results on synthetic datasets. Figure \ref{fig: feature selection visualization} provides detailed visualizations of results for the four priors with two representative signal levels $A=6$ and $A=10$ and with noise level $\sigma_\epsilon = 2$. Figure \ref{fig: feature selection MSE} summarizes the results for all the signal and noise levels. 
From Figure \ref{fig: feature selection MSE}, we see that  \texttt{BetaSS} with $p\sim \text{Beta}(2, 38)$ and \texttt{DeltaSS} with $p=0.05$ perform similarly when the noise level is low, but \texttt{DeltaSS} is better than  \texttt{BetaSS} for higher noise levels. This is because  the \texttt{DeltaSS} prior is more concentrated close to the true sparsity level; thus, it penalizes false signals more strongly (Top left and bottom panels in Figure \ref{fig: feature selection visualization}). The \texttt{InfoSS} prior has indistinguishable performance from the other \texttt{SS} priors when the noise level is low, but with high noise, e.g., $\sigma_\epsilon = 2.0$,  \texttt{InfoSS} is significantly better, especially when the signal is large ($A>6$). This is because \texttt{InfoSS} places a prior on the number of features directly, which makes the signals $w_i$ dependent, and consequently including correct signals can help remove incorrect signals.
In contrast, the signals are independent of each other in the other priors considered; thus, selecting true signals will not help remove false findings. Another observation is that the Horseshoe prior (\texttt{HS}) works well when there is little noise (e.g. $\sigma_\epsilon=1$), but for larger value of  $\sigma_\epsilon$ \texttt{HS} is much worse than all the spike-and-slab alternatives because it can easily overfit the noise (bottom-left panel in Figure \ref{fig: feature selection visualization}).\\

\subsection{Public real-world UCI datasets}

\textbf{Setup:} We analyze 7 publicly available datasets\footnote{https://archive.ics.uci.edu/ml/index.php}: \textit{Bike sharing}, 
\textit{California housing prices}, 
\textit{Energy efficiency}, 
\textit{Concrete compressive strength}, 
\textit{Yacht hydrodynamics}, \textit{Boston housing}, and \textit{kin8nm dataset},
whose characteristics, the number of individuals $N$ and the number of features $D$, are given in Table \ref{tab: UCI}.
We carry out two types of experiments: \underline{Original datasets}: we analyze the datasets as such, in which case there is no domain knowledge about sparsity; \underline{Extended datasets}: we concatenate 100 irrelevant features with the original features and add Gaussian noise to the dependent variable such that the PVE in the data is at most 0.2, which allows us to specify informative priors about sparsity (the number of relevant features is at most the number of original features) and the PVE (less than 0.2). We examine whether the performance can be improved by encoding this extra knowledge into the prior. We use $80\%$ of data for training and $20\%$ for testing. We use the PVE (i.e., $R^2$) on a test set to evaluate the performance\footnote{This is also consistent with the Mean Squared Error (MSE) when the residuals have zero mean.}. We repeat each experiment on $50$ different data splits to obtain confidence intervals.

\begin{table}[t]
\footnotesize
\caption{Details of the seven Gaussian scale mixture priors included in the comparison on the UCI datasets.  Hyper-parameters in \texttt{MF+CV} and \texttt{SS+CV} (local scale $\sigma_\lambda$ and spike probability $p$) are chosen via 5-fold cross-validation on the training set. The hyper-parameter $\beta$ is optimized to match the prior PVE, and $\mu_{+}$ is equal to the number of features in the corresponding original dataset without the artificially added irrelevant features. }
\label{tab: UCI model}
\begin{tabular}{c|c|c|c}
\hline
Name of the prior                   & $p(\lambda_i^{(l)}),\; \forall l\geq0$ & $p(\tau_i^{(0)})$ & $p(\tau_i^{(l)}),\; \forall l\geq1$ \\ \hline
\texttt{MF+CV}      & $\lambda_i^{(l)} = \sigma_\lambda$ & NA &  NA          \\ \hline
\texttt{SS+CV}     & $\lambda_i^{(l)} = \sigma_\lambda$ & Bernoulli($p$) &  Bernoulli($p$)           \\ \hline
\texttt{HS} & $C^{+}(0,1)$ & $p(\tau^{(0)})= C^{+}(0,10^{-5})$ & $p(\tau^{(l)})= C^{+}(0,10^{-5})$         \\ \hline
\texttt{HMF}             & vague Inv-Gamma &    NA    &  NA    \\ \hline
\texttt{InfoHMF}            & vague Inv-Gamma &    FL($0,D,1$)   &  NA     \\ \hline
\texttt{HMF+PVE}           & Inv-Gamma($2,\beta$) &    NA   &  NA     \\ \hline
\texttt{InfoHMF+PVE}          & Inv-Gamma($2,\beta$) &    FL($0,D,1$)  &  NA      \\ \hline
\end{tabular}
\end{table}
\begin{table}[t]
\centering
\scriptsize
\caption[Caption for LOF]{Test PVE with $1.96$ standard error of the mean (in parentheses) for each prior on UCI datasets. The first seven rows show the experimental results on the original datasets where we have no prior information, and the last seven rows on extended datasets with $100$ irrelevant features and injected noise added. The best result in each column has been boldfaced. The dimension ($P$)\footnotemark and size ($N$) are shown for each dataset. We see that both information about sparsity and PVE improve the performance, especially when prior information is available (on extended datasets).}
\label{tab: UCI}
\begin{tabular}{c|c|c|c|c|c|c|c}
\hline
\makecell{Original\\(P, N)}
& \makecell{California\\(\textbf{9}, 20k)}             & \makecell{Bike\\(\textbf{13}, 17k)}                   & \makecell{Concrete\\(\textbf{8}, 1k)}               & \makecell{Energy\\(\textbf{8}, 0.7k)}                 & \makecell{Kin8nm\\(\textbf{8}, 8.1k)}                 & \makecell{Yacht\\(\textbf{6}, 0.3k)}                  & \makecell{Boston\\(\textbf{3}, 0.5k)}                 \\ \hline
\texttt{Lasso+CV}               &0.629       &0.387       &0.631     &0.890        &0.890        &0.649        &0.666         \\ \hline
\texttt{MF+CV}               &\makecell{0.785\\(0.001)}        &\makecell{0.930\\(0.001)}        &\makecell{0.851\\(0.006)}        &\makecell{0.875\\(0.000)}        &\makecell{0.914\\(0.001)}        &\makecell{0.963\\(0.001)}        &\makecell{0.789\\(0.002)}         \\ 
\texttt{SS+CV}                &\makecell{0.785\\(0.001)}        &\makecell{0.925\\(0.001)}        &\makecell{0.875\\(0.003)}        &\makecell{0.893\\(0.001)}        &\makecell{0.908\\(0.001)}        &\makecell{0.924\\(0.004)}        &\makecell{0.804\\(0.003)}         \\ 
\texttt{HS}               &\makecell{0.795\\(0.001)}        &\makecell{0.925\\(0.002)}        &\makecell{0.859\\(0.003)}        &\makecell{0.897\\(0.002)}        &\makecell{0.905\\(0.001)}        &\makecell{0.944\\(0.003)}        &\makecell{\textbf{0.815}\\\textbf{(0.006)}}\\ 
\texttt{HMF}                   &\makecell{0.794\\(0.002)}        &\makecell{\textbf{0.932}\\\textbf{(0.001)}}&\makecell{0.895\\(0.002)}        &\makecell{0.960\\(0.005)}        &\makecell{0.930\\(0.001)}        &\makecell{\textbf{0.990}\\\textbf{(0.001)}}&\makecell{0.788\\(0.002)}         \\ 
\makecell{\texttt{HMF}\\\texttt{+PVE}}            &\makecell{\textbf{0.798}\\\textbf{(0.002)}}&\makecell{\textbf{0.932}\\\textbf{(0.001)}}&\makecell{\textbf{0.898}\\\textbf{(0.002)}}&\makecell{\textbf{0.967}\\\textbf{(0.001)}}&\makecell{\textbf{0.931}\\\textbf{(0.001)}}&\makecell{\textbf{0.990}\\\textbf{(0.001)}}&\makecell{0.801\\(0.003)}         \\ 
\texttt{InfoHMF}     &\makecell{0.795\\(0.002)}        &\makecell{\textbf{0.932}\\\textbf{(0.001)}}&\makecell{0.894\\(0.003)}        &\makecell{0.957\\(0.002)}        &\makecell{0.930\\(0.001)}        &\makecell{0.885\\(0.001)}        &\makecell{0.800\\(0.002)}         \\ 
\makecell{\texttt{InfoHMF}\\\texttt{+PVE}} &\makecell{0.797\\(0.002)}        &\makecell{\textbf{0.932}\\\textbf{(0.001)}}&\makecell{\textbf{0.898}\\\textbf{(0.002)}}&\makecell{0.961\\(0.002)}        &\makecell{0.930\\(0.001)}        &\makecell{0.989\\(0.002)}        &\makecell{0.804\\(0.002)}         \\ \hline
\makecell{Extended \\(P, N)}
& \makecell{California\\(\textbf{109}, 20k)}             & \makecell{Bike\\(\textbf{113}, 17k)}                   & \makecell{Concrete\\(\textbf{108}, 1k)}               & \makecell{Energy\\(\textbf{108}, 0.7k)}                 & \makecell{Kin8nm\\(\textbf{108}, 8.1k)}                 & \makecell{Yacht\\(\textbf{106}, 0.3k)}                  & \makecell{Boston\\(\textbf{103}, 0.5k)}                 \\ \hline
\texttt{Lasso+CV}               &0.131       &0.063       &0.101     &0.152        &0.075        &0.041        &0.025         \\ \hline
\texttt{MF+CV}                 &\makecell{ 0.064\\(0.004)}          &\makecell{ -0.029\\(0.007)}         &\makecell{ 0.000\\(0.000)}          &\makecell{ 0.003\\(0.003)}          &\makecell{ 0.010\\(0.007)}          &\makecell{ 0.002\\(0.003)}          &\makecell{ -0.002\\(0.005)}         \\ 
\texttt{SS+CV}                &\makecell{ 0.130\\(0.004)}          &\makecell{ 0.085\\(0.004)}          &\makecell{ 0.000\\(0.000)}          &\makecell{ 0.007\\(0.009)}          &\makecell{ 0.078\\(0.006)}          &\makecell{ 0.004\\(0.003)}          &\makecell{ 0.000\\(0.000)}          \\ 
\texttt{HS}                 &\makecell{ 0.046\\(0.006)}          &\makecell{ 0.001\\(0.009)}          &\makecell{ 0.111\\(0.020)}          &\makecell{ 0.155\\(0.022)}          &\makecell{ 0.031\\(0.009)}          &\makecell{ 0.163\\(0.022)}          &\makecell{ \textbf{0.105}\\\textbf{(0.022)}} \\ 
\texttt{HMF}                   &\makecell{ 0.138\\(0.004)}          &\makecell{ 0.149\\(0.004)}          &\makecell{ 0.081\\(0.024)}          &\makecell{ 0.148\\(0.019)}          &\makecell{ 0.132\\(0.006)}          &\makecell{ 0.112\\(0.035)}          &\makecell{ 0.052\\(0.019)}          \\ 
\makecell{\texttt{HMF}\\\texttt{+PVE}}                 &\makecell{ 0.139\\(0.003)}          &\makecell{ 0.151\\(0.004)}          &\makecell{ 0.106\\(0.019)}          &\makecell{ \textbf{0.160}\\\textbf{(0.020)}} &\makecell{ 0.136\\(0.006)}          &\makecell{ 0.138\\(0.039)}          &\makecell{ 0.075\\(0.021)}          \\ 
\texttt{InfoHMF}       &\makecell{ 0.139\\(0.01)}          &\makecell{ 0.159\\(0.005)}          &\makecell{ 0.092\\(0.024)}          &\makecell{ 0.148\\(0.020)}          &\makecell{ 0.150\\(0.006)}          &\makecell{ 0.131\\(0.030)}          &\makecell{ 0.058\\(0.019)}          \\ 
\makecell{\texttt{InfoHMF}\\\texttt{+PVE}} &\makecell{ \textbf{0.141}\\\textbf{(0.003)}} &\makecell{ \textbf{0.166}\\\textbf{(0.004)}} &\makecell{ \textbf{0.116}\\\textbf{(0.017)}} &\makecell{ 0.157\\(0.017)}          &\makecell{ \textbf{0.153}\\\textbf{(0.005)}} &\makecell{ \textbf{0.186}\\\textbf{(0.027)}} &\makecell{ 0.101\\(0.020)}          \\ \hline
\end{tabular}
\end{table}
\noindent\textbf{Parameter settings:} We considered 7 different priors: 1. mean-field (independent) Gaussian prior with cross-validation (\texttt{MF+CV}) \citep{blundell2015weight}; 2. delta spike-and-slab prior with cross-validation (\texttt{SS+CV}) \citep{blundell2015weight}; 3. horseshoe prior (\texttt{HS}) \citep{ghosh2017model}; 4. Hierarchical Gaussian prior (\texttt{HMF}), which is the same as \texttt{MF+CV} except that the hyperparameters receive a fully Bayesian treatment instead of cross-validation.
5. The \texttt{InfoHMF}, which incorporates domain knowledge on feature sparsity in the \texttt{HMF} by applying the proposed informative prior in the input layer; 6. \texttt{HMF+PVE} instead includes the informative prior on the PVE, and finally, 7. \texttt{InfoHMF+PVE} includes both types of domain knowledge. Lasso regression with cross-validated regularization (\texttt{Lasso+CV}) is included as another standard baseline \citep{tibshirani2011regression}.

The hyper-parameters for \texttt{MF+CV} priors and \texttt{SS+CV} prior are chosen by 5-fold cross-validation on the training set from grids $\sigma_\lambda\in\{e^{-2}, e^{-1}, 1, e^{1}, e^{2}\}$ and $p\in\{0.1, 0.3, 0.5,$ $0.7, 0.9\}$, which are wider than used in the original work by \cite{blundell2015weight}. We define \texttt{HS} as suggested by \cite{ghosh2018structured}, such that the scale $\tau_i^{(l)}=\tau^{(l)}$ is shared by all weights in each layer $l$. In the \texttt{HMF}, we use a non-informative prior on the local scales $\lambda_i^{(l)}$. We regard \texttt{MF+CV}, \texttt{SS+CV}, \texttt{HS}, and \texttt{HMF} as strong benchmarks to compare our novel informative priors against. For \texttt{InfoHMF}, we use the 'flattened' Laplace (FL) prior with $\mu_{-}=0, \mu_{+}=D$ ($D$ is the number of features in the original dataset) on the input layer to encode the prior knowledge about feature sparsity. For \texttt{HMF+PVE} and \texttt{InfoHMF+PVE}, we optimize the hyper-parameter $\beta$ to match the PVE of the BNN with a Beta$(5.0, 1.2)$ for the original datasets (the mode equals 0.95), and with Beta$(1.5, 3.0)$ for the extended datasets  (the mode equals 0.20). For all priors that are not informative about the PVE (except the HS), we use an Inv-Gamma(0.001,0.001) for the all layer-wise global scales $\boldsymbol{\sigma}$ and the noise $\sigma_\epsilon$. For priors informative on the PVE, the non-informative prior is used only for the last layer-wise global scale $\sigma^{(L)}$ and noise $\sigma_\epsilon$ (see Section \ref{sec: PVE}). The details about each prior are summarized in Table  \ref{tab: UCI model}.

\noindent\textbf{Results:} 
\footnotetext{The dimension $P=D$ in the original datasets, while $P=100+D$ in the extended datasets.}
The results, in terms of test PVE, are reported in Table \ref{tab: UCI}. For the \underline{original} \underline{datasets}, we see that incorporating the prior knowledge on the PVE (\texttt{HMF+PVE} and \texttt{InfoHMF+PVE}) always yields at least as good performance as the corresponding prior without this knowledge (\texttt{HMF} and \texttt{InfoHMF}). Indeed, \texttt{HMF+PVE} has the (shared) highest accuracy in all datasets except Boston. The new proposed informative sparsity inducing prior (\texttt{InfoHMF}) does not here improve the performance, as we do not have prior knowledge on sparsity in the original datasets. Among the non-informative priors, \texttt{HMF} is slightly better than the rest, except for the Boston housing dataset, where the horseshoe prior (\texttt{HS}) achieves the highest test PVE, which demonstrates the benefit of the fully Bayesian treatment vs. cross-validation of the hyperparameters. The linear method, \texttt{Lasso+CV}, is worse than all BNNs in most datasets.  

In the \underline{extended datasets} with the 100 extra irrelevant features and noise added to the target, knowledge on both the PVE and sparsity improves performance significantly. For most of the datasets both types of prior knowledge are useful, and consequently \texttt{InfoHMF+PVE} is the most accurate on 5 out of 7 datasets. Furthermore, its PVE is also close to 20\% of the maximum test PVE in the corresponding original dataset, reflecting the fact that noise was injected to keep only 20\% of the signal. We find that the horseshoe (\texttt{HS}) works better than the \texttt{HMF} on small datasets, especially Boston, where the \texttt{HS} outperform others. The priors \texttt{MF+CV} and \texttt{SS+CV} do not work well for the extended datasets, and they are even worse than \texttt{Lasso+CV}, because cross-validation has a large variance on the noisy datasets especially for flexible models such as BNNs. The more computationally intensive repeated cross-validation \citep{kuhn2013applied} might alleviate the problem, but its further exploration is left for future work. Overall, we conclude that by incorporating knowledge on the PVE and sparsity into the prior the performance can be improved; however, the amount of improvement can be small if the dataset is large (California and Bike) or when the prior knowledge is weak (the original datasets). 

\subsection{Metabolite prediction using genetic data}
\noindent\textbf{Setup:}
Genome-wide association studies (GWASs) aim to learn associations between genetic variants called SNPs (input features) and phenotypes (targets). Ultimately, the goal is to predict a given phenotype given the SNPs of an individual. This task is extremely challenging because 1. the input features are very high-dimensional and strongly correlated and 2. the features may explain only a tiny fraction of the variance of the phenotype, e.g. less $1\%$. In such a case, neural networks may overfit severly and have worse accuracy than the simple prediction by mean. Typical approaches employ several heuristic but crucial preprocessing steps to reduce the input dimension and correlation. However, strong domain knowledge about sparsity and the amount of variance explained by the SNPs is available, and we show that by incorporating this knowledge in the informative prior we can accurately predict where alternatives fail. 

We apply the proposed approach on the FINRISK dataset \citep{borodulin2018cohort}, which contains genetic data and 228 different metabolites as phenotypes for 4,620 individuals. We select six genes that have previously been found to be associated with the metabolites \citep{kettunen2016genome}. We use the SNPs in each gene as features to predict the metabolite most strongly associated with the gene, resulting in 6 different experiments. We make predictions using the posterior mean and evaluate the performance by the PVE (larger is better) on test data. We use 50\% of the data for training and 50\% for testing, and we repeat this 50 times for each of the six experiments (i.e., for each gene), to account for the variability due to BNN training.
\begin{table}[t]
\caption{The six Gaussian scale mixture priors included in the genetics experiment.  Hyper-parameters: the local scale $\sigma_\lambda$ and the spike probability $p$ are chosen via 5-fold cross-validation on the training set; the informative local scale $\sigma_\lambda^{\text{prior}}$ is optimized to match the prior PVE; $\mu_{+}$ equals $20\%$ number of SNPs in corresponding gene. }
\label{tab: FINRISK model}
\begin{tabular}{c|c|c|c}
\hline
Name of the prior                   & $p(\lambda_i^{(l)}),\; \forall l\geq0$ & $p(\tau_i^{(0)})$ & $p(\tau_i^{(l)}),\; \forall l\geq1$ \\ \hline
\texttt{MF+CV}     & $\lambda_i^{(l)} = \sigma_\lambda$ & NA &  NA          \\ \hline
\texttt{MF+PVE}           & $\lambda_i^{(l)} = \sigma_\lambda^{\text{prior}}$ & NA &  NA     \\ \hline
\texttt{SS+CV}    & $\lambda_i^{(l)} = \sigma_\lambda$ & Bernoulli($p$) &  Bernoulli($p$)           \\ \hline
\texttt{SS+PVE}    & $\lambda_i^{(l)} = \sigma_\lambda^{\text{prior}}$ & Bernoulli($p$) &  Bernoulli($p$)           \\ \hline
\texttt{InfoMF+CV}           & $\lambda_i^{(l)} = \sigma_\lambda$ &    FL($0,\mu_{\text{+}},1$)   &  NA     \\ \hline
\texttt{InfoMF+PVE}           & $\lambda_i^{(l)} = \sigma_\lambda^{\text{prior}}$ &    FL($0,\mu_{\text{+}},1$)  &  NA      \\ \hline
\end{tabular}
\end{table}

\noindent\textbf{Parameter settings:} We train BNNs with 1 hidden layer having $100$ hidden nodes; the complexity of the data prevents the use of more complex models, but even the single hidden layer increases flexibility and improves accuracy (see results). We consider six priors: the mean-field Gaussian with the local scales $\lambda_i^{(l)}$ set by cross-validation (\texttt{MF+CV}) or optimized using the PVE (\texttt{MF+PVE}); the delta spike-and-slab, where the slab probability $p$ is cross-validated and the local scales are set either by cross-validation or the PVE (\texttt{SS+CV} and \texttt{SS+PVE}); and finally, the mean-field prior including the informative prior about feature sparsity, and the local scales either cross-validated (\texttt{InfoMF+CV}) or set using the PVE (\texttt{InfoMF+PVE}). We use the `flattened' Laplace (FL) prior with $\mu_{-}=0, \mu_{+}=0.2D$, where $D$ is the number of SNPs in a given gene, to reflect the prior belief that less than $20\%$ of the SNPs in the gene affect the phenotype. To encode the knowledge of the PVE, we optimize the hyper-parameter $\sigma_\lambda$ to match the mode of the PVE with previous findings \citep{kettunen2016genome}. We also tried other priors such as the hierarchical Gaussian and the horseshoe but they failed to capture any signal even with prior knowledge on sparsity and the PVE, hence they are not included in the final comparison. The priors are summarized in Table  \ref{tab: FINRISK model}. We include Lasso regression with cross-validated regularization as a strong linear baseline \citep{lello2018accurate}.
\begin{figure*}[t]
    \centering
    \includegraphics[width=.95\linewidth]{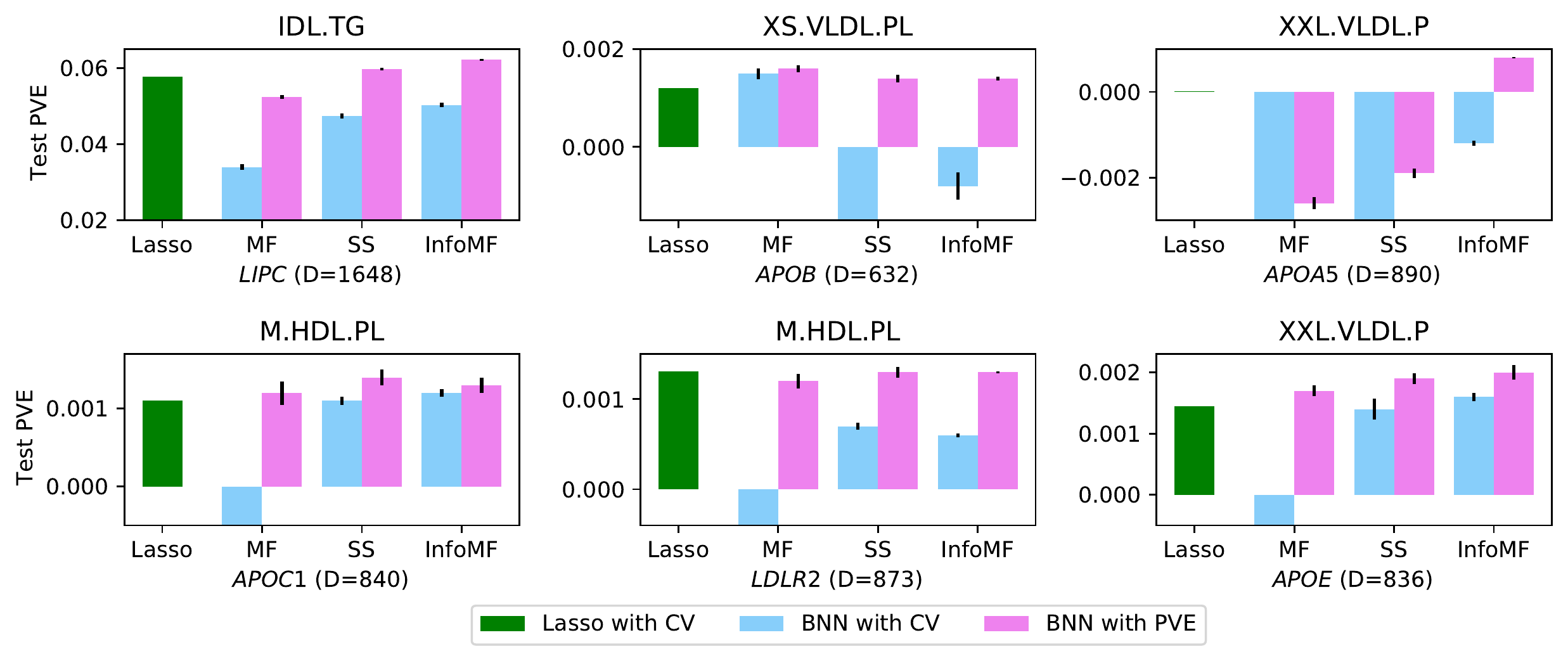}
    \caption{Each panel shows the results for one experiment of predicting a given metabolite (specified in the title, e.g., IDL.TG) using the SNPs in one gene (specified below the panel, e.g., \textit{LIPC}). Each bar shows the average test PVE over 50 repeated experiments, and the error bar is the corresponding 95\% CI. Blue bars indicate priors not including knowledge on the PVE (with cross-validated hyper-parameters), while purple bars show priors that incorporate the knowledge on the PVE. Green bars show Lasso linear regression with cross-validated regularization. Negative test PVE values represent overfitting. Some results that fall below the scale have been removed for illustration purposes. In summary, both information about the PVE and sparsity improve the performance in most experiments.
    }
    \label{fig: GWAS}
\end{figure*}

\noindent\textbf{Results:} Figure \ref{fig: GWAS} shows results for the 6 experiments (genes). We see that using the prior knowledge on the PVE always improves accuracy (purple bars). Without the prior on the PVE the mean-field Gaussian prior can overfit severly (blue bars, negative test PVE). Furthermore, the novel informative sparse prior performs better than or similarly to the spike-and-slab prior with the cross-validated slab probability (compare \texttt{InfoMF} vs. \texttt{SS}). However, it is notable that applying the \texttt{SS} prior requires computationally intensive cross-validation to set the slab probability $p$, which is avoided by the \texttt{InfoMF}. The gene \textit{LIPC} has the strongest signal and the BNNs can perform well even without the informative priors, but the performance improves further by incorporating the prior knowledge. For gene \textit{APOA5}, only \texttt{InfoMF+PVE} achieves a positive test PVE, while the alternatives overfit severly. We also notice that although BNNs with informative priors are better than the Lasso in most cases, the performance is similar for gene \textit{LDLR2}, which indicates that the true effect can be captured by a linear model. Overall, the highest accuracy is achieved by \texttt{SS+PVE} or \texttt{InfoMF+PVE} priors in most experiments.

\section{Conclusion}\label{sec:conclusion}
In this paper, we provided an approach to incorporate two types of domain knowledge, on feature sparsity and the proportion of variance explained (PVE), into the widely used Gaussian scale mixture priors for Bayesian neural networks. Specifically, we proposed to use a new informative spike-and-slab prior on the input layer to reflect the belief about feature sparsity, and to tune the model's PVE with prior knowledge on the PVE, by optimizing the hyper-parameters of the local scales for all neural network weights. We demonstrated the utility of the approach on simulated data, publicly available datasets, and in a genetics application, where they outperformed strong commonly used baselines without computationally expensive cross-validation. The informative spike-and-slab is not limited to the Gaussian scale mixtures, but can be generalized to all scale mixture distributions. However, one limitation of using the PVE to reflect the signal-to-noise ratio is that PVE is only defined for priors with finite second moments. Therefore, for some heavy-tailed distributions, such as the horseshoe, other measures of signal-to-noise ratio should be developed as part of future work.

\begin{supplement}
Supplementary material of ``Informative Bayesian Neural Network Priors forWeak Signals" contains proofs and implementation details
\end{supplement}

\bibliographystyle{ba}
\bibliography{informative_prior}

\begin{acknowledgement}
The work used computer resources of the Aalto University School of Science Science-IT project. This work was supported by the Academy of Finland (Flagship programme: Finnish Center for Artificial Intelligence, FCAI, and grants 319264, 292334, 286607, 294015).
\end{acknowledgement}

\end{document}